\documentclass[conference]{IEEEtran}
\IEEEoverridecommandlockouts
\usepackage{cite}
\usepackage{amsmath,amssymb,amsfonts}
\usepackage{algorithmic}
\usepackage{graphicx}
\usepackage{textcomp}
\usepackage{xcolor}
\usepackage{makecell}
\usepackage{multicol}
\usepackage{multirow}
\usepackage{bm}
\usepackage{booktabs}
\usepackage{svg}
\usepackage{colortbl}
\usepackage{subfig}
\usepackage{subfloat}
\usepackage{mathtools}
\usepackage{hyperref}

\newcommand{\methodname}{SciTopic}
 
\def\BibTeX{{\rm B\kern-.05em{\sc i\kern-.025em b}\kern-.08em
    T\kern-.1667em\lower.7ex\hbox{E}\kern-.125emX}}
\begin{document}

\title{SciTopic: Enhancing Topic Discovery in Scientific Literature through Advanced LLM
}

\author{\IEEEauthorblockN{Pengjiang Li$^{1,2,\textsuperscript{\textdagger}}$, Zaitian Wang$^{1,2,\textsuperscript{\textdagger}}$ \thanks{\textsuperscript{\textdagger}These authors contributed equally to this work.}, Xinhao Zhang$^{3}$, Ran Zhang$^{1,2}$, Lu Jiang$^{4}$,  Pengfei Wang$^{1,2,*}$\thanks{$^*$ Corresponding author.}, Yuanchun Zhou$^{1,2}$
}
\IEEEauthorblockA{\textit{$^1$Computer Network Information Center, Chinese Academy of Sciences, Beijing, China} \\
\textit{$^2$University of Chinese Academy of Sciences, Beijing, China} \\
\textit{$^3$Department of Computer Science, Portland State University, Portland, US} \\
\textit{$^4$Information Science and Technology College, Dalian Maritime University, Dalian, China} \\
\{pjli,zhangran,zyc\}@cnic.cn, wangzaitian23@mails.ucas.ac.cn, xinhaoz@pdx.edu, jiangl761@dlmu.edu.cn, wpf2106@gmail.com
}}



\maketitle

\begin{abstract}
Topic discovery in scientific literature provides valuable insights for researchers to identify emerging trends and explore new avenues for investigation, facilitating easier scientific information retrieval. 
Many machine learning methods, particularly deep embedding techniques, have been applied to discover research topics. 
However, most existing topic discovery methods rely on word embedding to capture the semantics and lack a comprehensive understanding of scientific publications, struggling with complex, high-dimensional text relationships. 
Inspired by the exceptional comprehension of textual information by large language models (LLMs), we propose an advanced topic discovery method enhanced by LLMs to improve scientific topic identification, namely \textbf{SciTopic}. 
Specifically, we first build a textual encoder to capture the content from scientific publications, including metadata, title, and abstract.
Next, we construct a space optimization module that integrates entropy-based sampling and triplet tasks guided by LLMs, enhancing the focus on thematic relevance and contextual intricacies between ambiguous instances. 
Then, we propose to fine-tune the textual encoder based on the guidance from the LLMs by optimizing the contrastive loss of the triplets, forcing the text encoder to better discriminate instances of different topics. 
Finally, extensive experiments conducted on three real-world datasets of scientific publications demonstrate that \textbf{SciTopic} outperforms the state-of-the-art (SOTA) scientific topic discovery methods, enabling researchers to gain deeper and faster insights\footnote{
Access the source code link: 
{\color{black}\hyperlink{https://github.com/CNICDS/SciTopic}{https://github.com/CNICDS/SciTopic}}}. 

\if false
The exponential growth of scientific publications, particularly in computer science, presents substantial challenges for researchers attempting to navigate this vast sea of information. 
In response, we propose an advanced topic modeling framework named \textbf{SciTopic} that employs large language models (LLMs) and sophisticated document embeddings to improve topic identification. 
Our approach distinctively avoids the use of dimensionality reduction techniques, thereby maintaining the complex high-dimensional structure of textual data and capturing subtle thematic relationships more effectively. 
This method integrates entropy-based sampling and triplet tasks to refine the clustering process, enhancing focus on thematic relevance and contextual intricacies. 
By fine-tuning the embedding model, we significantly enhance its capability to identify and delineate nuanced topic structures. 
We validated our framework with datasets from prominent Artificial Intelligence (AI) and Data Mining (DM) conferences, where it demonstrated a marked improvement in detecting complex topic dynamics over traditional models. 
The findings underscore the potential of our framework to substantially advance the analysis of scientific literature, rendering the review process more precise and insightful for researchers.
\fi

\begin{IEEEkeywords}
scientific topic discovery, text clustering, large language models, document embeddings
\end{IEEEkeywords}

\end{abstract}
\section{Introduction}
As the frontiers of science continue to expand, scholars are inundated with an ever-growing influx of information disseminated across numerous scientific publications. 
The proliferation of scientific literature, particularly in rapidly evolving fields like computer science, poses significant challenges to information retrieval and management, making it increasingly difficult to stay abreast of the latest developments. 
Topic discovery serves as a foundational element in facilitating scientific information retrieval, enabling researchers to navigate the complexities of their disciplines with greater ease and precision. 
Traditional information retrieval methods, relying on manual curation or basic keyword searches, often fail to capture the nuanced relationships between different research areas or overlook emerging interdisciplinary connections. 
In response, automated scientific topic discovery is urgently needed to effectively handle the increasing complexity and scale of modern scientific literature. 

Recent advancements in machine learning, particularly in the realm of deep learning~\cite{xu2025scsiameseclu,wang2024deep,zhang2025motif}, have led to the emergence of various techniques aimed at automating the topic discovery process. 
The classical topic modeling techniques, including Latent Dirichlet Allocation (LDA)~\cite{blei2003latent}, Non-negative Matrix Factorization (NMF)~\cite{lee2000algorithms}, and Probabilistic Latent Semantic Analysis (PLSA)~\cite{hofmann1999probabilistic}, could be applied to discover the scientific topics directly.
However, these bag-of-words methods ignore contextual word interrelations, failing to capture the intricate semantics of modern scientific texts. 
In addition, these techniques often necessitate dimensionality reduction processes such as Principal Component Analysis (PCA)~\cite{abdi2010principal} or Uniform Manifold Approximation and Projection (UMAP)~\cite{mcinnes2018umap}, potentially leading to significant information loss that is vital for maintaining the thematic depth of the documents. 
Deep embedding methods have gained prominence for their ability to represent textual data in a high-dimensional space, capturing semantic relationships between words and phrases. 
Unlike traditional bag-of-words approaches that treat words in isolation, document embeddings encapsulate the overall significance of a document in a continuous vector space, aligning semantically akin words more closely~\cite{mikolov2013efficient}. 
Even advanced deep topic modeling like the Embedded Topic Model (ETM)~\cite{dieng2020topic} and Neural Variational Document Model (NVDM)~\cite{miao2016neural}, though incorporating word embeddings to capture semantic nuances, 
still results in a limited understanding of the intricate relationships. 
These limitations ultimately impact the quality of insights derived from topic discovery, potentially leading to incomplete or inaccurate representations of the underlying thematic structures within the literature.

To overcome these limitations, we draw inspiration from the remarkable capabilities of large language models (LLMs) in comprehending textual information. 
LLMs, such as GPT-4 and BERT, have exhibited unparalleled proficiency in natural language comprehension by recognizing deep contextual connections between words, phrases, and entire texts. 
Transformer-based architectures can capture long-range dependencies and process substantial volumes of text in a context-sensitive manner~\cite{vaswani2017attention}, effectively discerning nuanced thematic structures. 
Trained on vast data, transformer-based models can capture complex patterns, contextual cues, and semantic relationships beyond mere word co-occurrence. 
These embeddings retain syntactic and semantic linkages, offering comprehensive depictions of documents’ thematic content compared to earlier sparse models. 
This enhancement allows for refined topic modeling and improved semantic clustering, naturally grouping documents based on their intrinsic meanings rather than mere word frequency. 
For instance, Sentence-BERT (SBERT) optimizes BERT embeddings for semantic similarity tasks using a siamese network structure, preserving intricate semantic details and capturing subtle thematic distinctions~\cite{reimers2019sentence}. 

Along this line, by leveraging the strengths of LLMs, we propose SciTopic, an effective method to enhance scientific topic identification and provide deeper insights for researchers. 
Firstly, we construct a text encoder that captures essential content from scientific publications, including metadata, titles, and abstracts. 
Through this module, we can extract meaningful textual features crucial for accurate topic identification. 
Then, we introduce an LLM-guided clustering technique that leverages entropy-based sampling and triplet tasks. 
This approach diverges from traditional unsupervised clustering methods by actively involving the LLM in the clustering process. 
We utilize an entropy-based sampling strategy to identify the most ambiguous or uncertain documents, where cluster membership is less defined. 
These high-entropy instances are used as anchors for the triplet tasks, where two candidate titles or abstracts from nearby clusters are selected. 
By analyzing these triplets, the LLM refines document embeddings, sharpening the distinctions between closely related clusters. 
This method not only improves clustering precision but also minimizes computational overhead by focusing the LLM's attention toward the most informative cases. 
Ultimately, this approach leads to a more contextually accurate and thematically coherent clustering result, ensuring that even subtle topic differences are effectively captured. 
Our key contributions are as follows:

\vspace{-1mm}
\begin{itemize}
    \item We propose a novel and comprehensive topic discovery framework enhanced by LLM-guided clustering and entropy-based sampling, which effectively refines document embeddings and strategically focuses on the most ambiguous and uncertain cases, thereby significantly improving the overall accuracy, robustness, and thematic coherence of topic discovery.
    \item We design a prompt-based triplet task for LLMs to differentiate closely related scientific documents, and utilize the generated responses to fine-tune embedding models for more distinctive representations. 
    \item We construct extensive experiments on real-world scientific literature datasets, including a dataset specifically curated for this study, to demonstrate the superior performance of the proposed model, \textbf{SciTopic}, which consistently outperforms state-of-the-art methods across topic and clustering evaluation metrics. 
\end{itemize}

\section{Related Work}

\subsection{Neural Topic Models}
Neural Topic Models (NTMs) integrate deep learning with probabilistic modeling to improve topic coherence and document representation~\cite{lee2000algorithms,zhao2020neural,dieng2020topic,Zhao2021TopicMM,wu2022mitigating, yang-etal-2025-neural}. LDA inspires many neural adaptations addressing scalability and coherence for large vocabularies~\cite{blei2003latent}. Recent advancements include ECRTM, which prevents topic collapse through embedding clustering regularization~\cite{wu2023effective}, and FASTopic, utilizing dual semantic-relation reconstruction to regulate embeddings~\cite{wu2024fastopic}. InfoCTM aligns topics cross-lingually using mutual information~\cite{wu2023infoctm}, while BERTopic leverages pre-trained transformers and clustering for coherent topics~\cite{grootendorst2022bertopic}. Despite these advancements, most NTMs are not tailored for scientific topic discovery tasks.

\subsection{Scientific Topic Discovery}
Scientific topic discovery automates the identification of research trends and emerging areas, essential given the rapid growth in publications~\cite{langley1987scientific,Huang2022TopicDI,ninkov2022bibliometrics}. 
Domain-specific models assess research impact~\cite{Guo2014ATT,Zhang2018DomainspecificTM}. 
For instance, ~\cite{PonsPorrata2007TopicDB} adapts hierarchical topic discovery for scientific literature, and ~\cite{Jeong2016DiscoveryOR} tracks authors' research evolution. ~\cite{Altaf2021ScientificDD} proposes dataset recommendation at the topic level, while ~\cite{Wu2024EmergingST} identifies emerging topics by analyzing rare synonymous biterms. 
However, most methods fail to adequately capture complex relationships. 

\subsection{Prompt Learning}
Prompt learning advances LLM applications by designing task-specific cues, enabling few-shot and zero-shot scenarios~\cite{Liu2021PretrainPA,He2023YouOP}. Techniques like domain-controlled prompts for remote sensing~\cite{Cao2023DomainControlledPL}, graph prompt learning~\cite{Sun2023GraphPL}, and Match-Prompt frameworks for diverse tasks~\cite{Xu2022MatchPromptIM} have demonstrated effectiveness. Methods such as region-based image recognition~\cite{Zhou2023HeterogeneousRE}, unsupervised image enhancement~\cite{Liang2023IterativePL}, and continual learning~\cite{Tang2023WhenPI} further showcase prompt learning’s versatility. Inspired by prompt-based learning, we use prompts to improve query triplet generation for scientific topic discovery.

\section{Method}
\begin{figure*}[htbp]
    \centering
    \includegraphics[width=1.0\linewidth]{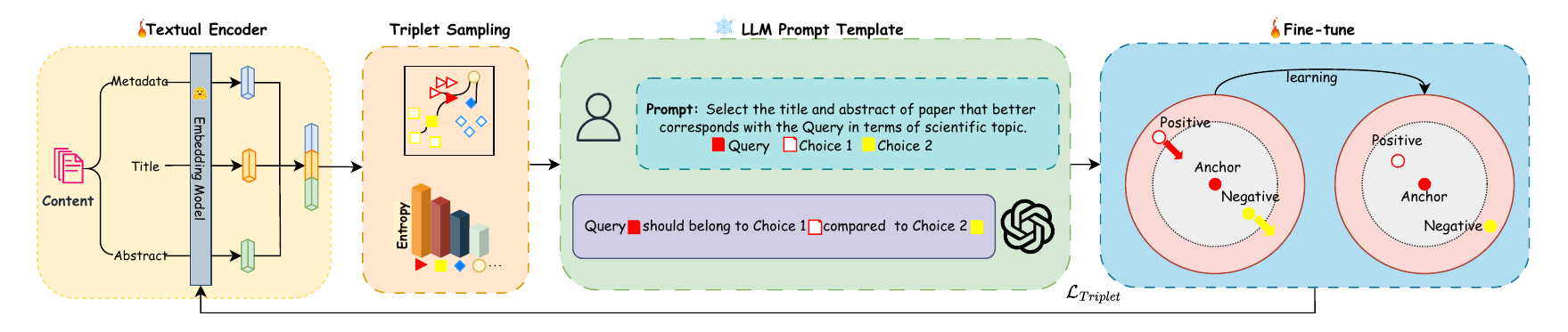}
    \caption{Overview of the proposed \textbf{SciTopic} framework. The framework comprises three key stages: a) Textual encoder, where title, abstract, and metadata are separately encoded and concatenated to form a comprehensive document representation; b) LLM-guided clustering, which leverages LLM-guided triplet tasks and entropy-based sampling to handle thematically ambiguous documents, enhancing clustering precision through LLM feedback; and c) Fine-tuning, where the embedding model is optimized using LLM triplet feedback to produce final clustering results with improved thematic relevance and coherence.}
    \label{fig:framework}
\end{figure*}

\subsection{Problem Definition}
In our task, we aim to extract meaningful research topics from large collections of scientific papers. 
Each paper is in the form of a document containing title ($t$), abstract ($a$), and metadata ($m$) and can be represented by a set of textual features ${\{x_1, x_2, ..., x_n\}}$. 
Then, we cluster documents into distinct groups ${\{C_1, C_2, ..., C_k\}}$. From each cluster, we further extract a list of key terms to verbalize the represented topic, enabling efficient topic identification and trend analysis. 

\subsection{Framework Overview}

As depicted in Figure~\ref{fig:framework}, our method starts with encoding the title, abstract, and metadata of each document and concatenating their embeddings to a composite feature matrix. Then we group these documents into clusters and sample ambiguous instances to construct the triplet task, which prompts the LLM to evaluate and reassign documents to more coherent clusters. These refined triplets are finally used with contrastive learning~\cite{ning2025rethinkinggraphcontrastivelearning} to fine-tune the embedding model for an enhanced clustering performance. 

\subsection{Textual Encoder}

To create effective representations of scientific documents that improve clustering and retrieval, we use a dynamic embedding framework. This framework processes text at various levels of detail, from single sentences to entire documents, and adapts well to diverse text types such as short titles, detailed abstracts, and full papers. It also allows for contextual and domain-specific fine-tuning to enhance performance.
Our approach generates document embeddings by separately encoding each document's title, abstract, and metadata. These embeddings are then combined into a unified representation. Metadata includes bibliographic details such as author list, publication year, and conference venue, adding contextual depth to each document's profile.

Each document is divided into three key components: the title, abstract, and metadata. The title and abstract are encoded separately using a fine-tuned model capable of handling varied text granularities. Metadata is consolidated into a single string and embedded using the same model to maintain consistency. The encoding process is expressed as follows:
\begin{equation}
\begin{aligned}
\bm{h}^t_j =& f(x^t_j) \\
\bm{h}^a_j =& f(x^a_j) \\
\bm{h}^m_j =& f(x^m_j)
\end{aligned}
\end{equation}
where $f$ represents the embedding model, and $x^t_j$, $x^a_j$, and $x^m_j$ denote the title, abstract, and metadata respectively. The embeddings are concatenated as a composite feature matrix:
\begin{equation}
\bm{h}^p_j = \text{Concat}(\bm{h}^t_j, \bm{h}^a_j, \bm{h}^m_j),
\end{equation}
where $\bm{h}^p_j$ is the final embedding for document $j$, integrating content (title and abstract) and context (metadata). This combined representation ensures a comprehensive understanding of each document.
By leveraging a fine-tunable model, our method effectively captures both semantic content and contextual information. This enhances clustering and retrieval performance across large and diverse datasets. The framework's adaptability allows it to meet specific dataset requirements while maintaining generalization across domains, making it suitable for a wide range of applications.

\subsection{LLM-Guided Clustering}

Here, we refine text clustering using a triplet task guided by large language models (LLMs), designed to reflect user-specific perspectives. 
Each triplet comprises an anchor and two candidate elements $(a, c_1, c_2)$, aiming to identify which candidate is more thematically similar to the anchor.

LLMs have emerged as powerful tools for refining text clustering by leveraging their advanced contextual understanding and adaptability. This section details how LLMs are utilized to perform guided tasks that improve clustering alignment with user-specific perspectives and thematic relevance.

\noindent\textbf{Entropy-Based Sampling for Efficient Clustering.}
To maximize efficiency, an entropy-based sampling method selects the most informative triplet, involving two key steps:

\noindent\textbf{\textit{Entropy Calculation for Ambiguous Instances.}} The entropy of each document embedding is calculated to identify high-uncertainty cases regarding cluster assignments. These high-entropy instances serve as anchors, representing the most ambiguous clustering scenarios. Using the K-Means algorithm, each document is linked to a cluster center, denoted as $\mu_i$. A probabilistic t-distribution mechanism is used to compute soft assignments, providing a nuanced understanding of the probability of a document belonging to each cluster:
\begin{equation}
P(\bm{\mu}_i | \bm{h}^p_j) = \frac{(1 + \frac{\Vert \bm{h}^p_j - \bm{\mu}_i \Vert^2}{\alpha})^{-\frac{\alpha + 1}{2}}}{\sum_k(1 + \frac{\Vert \bm{h}^p_j - \bm{\mu}_k \Vert^2}{\alpha})^{-\frac{\alpha + 1}{2}}},
\end{equation}
where $\Vert \bm{h}^p_j - \bm{\mu}_i \Vert^2$ represents the squared Euclidean distance between document embedding $\bm{h}^p_j$ and cluster centroid $\mu_i$, and $\alpha$ controls the t-distribution's degrees of freedom. Soft assignments allow for a more detailed and flexible representation of overlapping thematic areas.

To limit the cost of entropy computation, entropy is calculated only for a subset of nearby clusters, determined by:
\begin{equation}
\phi = \max(\lambda K, 2),
\end{equation}
where $\lambda$ is a scaling factor and $K$ is the total number of clusters. For these selected clusters, probabilities are normalized:
\begin{equation}
\hat{P}(\bm{\mu}_i | \bm{h}^p_j) = \frac{P(\bm{\mu}_i | \bm{h}^p_j)}{\sum_{i=1}^{\phi}P(\bm{\mu}_i | \bm{h}^p_j)}
\end{equation}
The entropy for paper $j$ is calculated as:
\begin{equation}
H(x_j) = -\sum^{\phi}_{i=1}\hat{P}(\bm{\mu}_i | \bm{h}^p_j)\log(\hat{P}(\bm{\mu}_i | \bm{h}^p_j))
\end{equation}
Entropy-guided bounding parameters, $\sigma_{low}$ and $\sigma_{high}$, regulate the number of clusters considered, effectively balancing computational efficiency and analytical depth. 

\noindent\textbf{\textit{Sampling from Closest Clusters.}} High-entropy anchors, representing ambiguous or uncertain instances, are paired with candidate points sampled from nearby clusters based on cosine similarity of their embeddings. Specifically, we identify the closest clusters by calculating the mean embedding vectors of all clusters and selecting clusters with minimal Euclidean or cosine distances to the anchor. To enhance informativeness, candidates are then sampled proportionally to the density of these nearby clusters to ensure diverse coverage.

This targeted sampling generates informative triplets \((\text{anchor}, \text{positive}, \text{negative})\), where the positive is sampled from the same cluster as the anchor, and the negative is sampled from a closely related but distinct cluster. 
By focusing on ambiguous instances, this entropy-driven method reduces LLM query frequency and improves cost-effectiveness while maintaining high-quality clustering results.

\noindent\textbf{Triplet Task for Clustering Perspective.} At the core of the method is a triplet task that allows LLMs to evaluate thematic relationships among three elements: an anchor and two candidates. The LLM is prompted with the query:
\begin{equation}
\rho = \text{``Select the paper closest to }~a:\ c_1,c_2,\text{or Neither.''} \nonumber
\end{equation}
The LLM processes this prompt and determines which candidate aligns more closely with the anchor in terms of thematic similarity. If the LLM responds with "Neither," the triplet is excluded, ensuring only meaningful comparisons guide the clustering process. This mechanism enhances contextual precision by eliminating ambiguous or irrelevant triplets, providing a cleaner input for fine-tuning the clustering model.

Each valid triplet consists of an anchor, a positive example (closer candidate), and a negative example (distant candidate), which are used to fine-tune the embedding model. By learning from these structured comparisons, the model develops embeddings that capture nuanced thematic distinctions, improving clustering coherence and accuracy.

\subsection{Fine-tuning}

To enhance the discriminative power of the embedding model, triplets generated by the LLM guide the fine-tuning process. Each triplet includes an anchor document, a thematically similar document (positive example), and a thematically different document (negative example): \(t = (a, c^{+}_i, c^{-}_i)\). These triplets are generated by analyzing high-entropy anchors, where uncertainty in the clustering assignments is highest. The LLM helps to refine thematic boundaries by sampling from nearest clusters, ensuring the selection of highly informative positive and negative examples.

The fine-tuning process employs a cross-entropy loss with in-batch negative sampling~\cite{wang2021cross}:
\begin{equation}
\mathcal{L} = -\log\frac{e^{s(a, c^{+}_i) / \tau}}{\sum_{c_j \in \mathcal{D}} e^{s(a, c_j) / \tau}},
\end{equation}
where \(s(x, y)\) denotes the similarity score (e.g., cosine similarity or dot product) between the embeddings of \(x\) and \(y\), \(\mathcal{D}\) represents the set of all positive and negative pairs in the current batch, and \(\tau\) is the softmax temperature that controls the sharpness of the probability distribution. A lower \(\tau\) sharpens the distribution, emphasizing the strongest positive-negative contrast, while a higher \(\tau\) creates a smoother distribution, which can be beneficial in noisy environments.

To further improve robustness, the similarity function \(s(x, y)\) incorporates margin-based adjustments:
\begin{equation}
s(x, y) = \frac{f(x) \cdot f(y)}{\|f(x)\| \|f(y)\|} - \gamma,
\end{equation}
where \(f(x)\) is embedding function, and \(\gamma\) is a margin parameter that helps avoid trivial solutions by enforcing a minimal semantic distinction between positive and negative pairs. 

The fine-tuning process iteratively updates the embedding model to create a more compact and semantically meaningful latent space. By emphasizing semantic distinctions between positive and negative samples, the refined embeddings result in improved clustering outcomes, particularly for high-entropy regions where thematic boundaries are less clear. To further enhance model adaptability, the batch construction strategy incorporates dynamic sampling, which increases the weight of high-uncertainty samples, ensuring the model efficiently learns from challenging instances.

\subsection{Topic Verbalization}
\begin{figure}[bp]
\centering
\includegraphics[width=1.0\linewidth]{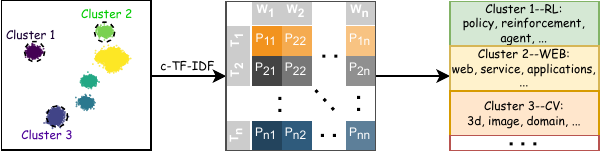}
\caption{Illustration of class-based TF-IDF analysis.}
\label{fig:topic-verbalization}
\end{figure}

As depicted in Figure~\ref{fig:topic-verbalization}, we use class-based term frequency-inverse document frequency (c-TF-IDF)~\cite{grootendorst2022bertopic}, which calculates scores for each cluster by treating it as a single document, emphasizing key terms that define each topic. The weight of a term \(t\) in cluster \(c\) is calculated as:
\begin{equation}
W^{'}_{t, c} = \left(\frac{tf_{t, c}}{T_{c}}\right) \cdot \log\left(1 + \frac{B}{cf_t}\right),
\end{equation}
where \(W^{'}_{t, c}\) represents the importance of term \(t\) in cluster \(c\), \(tf_{t, c}\) is the term frequency of \(t\) in cluster \(c\), \(T_{c}\) is the total term count in the cluster, \(B\) is the average term count across all clusters, and \(cf_t\) is the cluster frequency of \(t\) (i.e., the number of clusters containing term \(t\)). 
Compared to standard TF-IDF, c-TF-IDF adapts the term weighting to a cluster-level analysis by treating each cluster as a pseudo-document, enabling the identification of terms that are distinctive to each topic. 
This approach is particularly effective for unsupervised topic models, where clusters overlap semantically, as it highlights each cluster's unique characteristics and reduces the influence of common terms. 

To further enhance topic representation, the terms with the highest \(W^{'}_{t, c}\) values are selected to represent each cluster. 
These terms are ranked based on their contribution to cluster uniqueness, enabling a concise yet informative summary of each topic. 
Additionally, this representation can be visualized using word clouds
, where the font size of each term reflects its corresponding weight, offering an intuitive understanding of the most salient terms within each topic. 
Finally, c-TF-IDF allows the integration of thematic structure into downstream tasks, such as topic classification or document clustering. 
By capturing nuanced differences between clusters, this method significantly improves the interpretability and accuracy of topic modeling results.

\section{Experiment}
This section introduces the datasets used in this study, including two additional benchmark datasets. We evaluate our proposed method, \methodname, using topic and clustering metrics, followed by a thorough analysis of the semantic properties of the clustering results. Finally, we conduct an importance analysis of the model components and evaluate parameters.
\subsection{Datasets}
This study identifies prevalent topics in scholarly articles using the DBLP database and two additional Kaggle datasets, as shown in Table~\ref{tab:dataset}. The \textbf{AI-DM} Research Literature Dataset, based on the DBLP database, includes metadata for 57,320 AI papers and 20,700 DM papers from major conferences such as AAAI, ACL, CVPR, IJCAI, NeurIPS, and SIGKDD, covering titles, authors, publication years, and venues. 
Additionally, the \textbf{DBLP V10}\footnote{ {\color{black}\hyperlink{https://www.kaggle.com/datasets/nechbamohammed/research-papers-dataset}{https://www.kaggle.com/datasets/nechbamohammed/research-papers-dataset}}} Dataset, derived from DBLP (Version 10), spans fields like computer science, mathematics, and physics, offering metadata for 999,064 papers, including 827,533 with both titles and abstracts, from which 100,000 papers were randomly sampled. The \textbf{NeurIPS}\footnote{ {\color{black}\hyperlink{https://www.kaggle.com/datasets/benhamner/nips-papers}{https://www.kaggle.com/datasets/benhamner/nips-papers}}} Dataset provides detailed information on 7,241 NeurIPS conference papers from 1987 to 2016, including titles, authors, abstracts, and full texts, offering comprehensive insights into machine learning advancements. This combination of datasets ensures robust coverage of both general and domain-specific scholarly topics for analysis.
Furthermore, this study incorporates additional datasets for broader coverage, including 20,000 randomly sampled papers from the \textbf{arXiv}\footnote{ {\color{black}\hyperlink{https://www.kaggle.com/datasets/Cornell-University/arxiv}{https://www.kaggle.com/datasets/Cornell-University/arxiv}}}, and the \textbf{PubMed Dataset}\footnote{ {\color{black}\hyperlink{https://www.kaggle.com/datasets/nabarupghosh/pubmed-medical-dataset-2014-to-2023-title-abstract}{https://www.kaggle.com/datasets/nabarupghosh/pubmed-medical-dataset-2014-to-2023-title-abstract}}} containing 142,432 papers from 2014 to 2023 in the biomedical field. These diverse datasets ensure robust coverage of both general and domain-specific scholarly topics for analysis.

\begin{table}[!t]
    \centering
    \vspace{-2mm}
    \resizebox{\linewidth}{!}{
    
        \begin{tabular}{llc}
            \toprule
             Dataset & Conference/journal/subject & Paper number \\
            \midrule
            \multirow{1}{*}{NeurIPS} & NeurIPS & 7238\\
            \midrule
            \multirow{14}{*}{AI-DM} & AAAI & 9769 \\
                           & ACL & 3507 \\
                           & CVPR & 8038 \\
                           & ICCV & 3172 \\
                           & ICML & 4477 \\
                           & IJCAI & 5121 \\
                           & NIPS & 15899 \\
                           & WWW & 7337 \\
                           & SIGKDD & 5573 \\
                           & ICDM & 135 \\
                           & SIGIR & 5608 \\
                           & CIKM & 7765 \\
                           & SDM & 1619 \\
            \midrule
            \multirow{9}{*}{DBLP V10} & ICASSP & 11770 \\
            & ICRA & 9573 \\
            & LNCS & 7606 \\
            & IEEE ICC & 7272 \\
            & IROS & 6968 \\
            & ICIP & 6757 \\
            & GLOBECOM & 6651 \\
            & IGARSS & 6098 \\
            & Others & 49075\\
            \midrule
            \multirow{7}{*}{arXiv} & High Energy Physics - Phenomenology & 10100 \\
            & Computer Vision & 8746 \\
            & Quantum Physics & 8556 \\
            & High Energy Physics - Theory & 8034 \\
            & Machine Learning & 7527 \\
            & Astrophysics & 6890 \\
            & Others & 150147 \\
            \midrule
            \multirow{5}{*}{PubMed} & Sensors & 6773 \\
            & Scientific reports & 4700 \\
            & PloS one & 3028 \\
            & IEEE TNNLS & 1866 \\
            & Others & 126065 \\
            \bottomrule
        \end{tabular}
    }
    \caption{Statistics of the dataset.}
    \label{tab:dataset}
\end{table}

\begin{table*}[!t]
    \centering
    \setlength{\tabcolsep}{1.5mm}
    \resizebox{\linewidth}{!}
    {
    \begin{tabular}{lrrrrrrrrrrrrrrrrrrrrrrrr}
        \toprule
        \multirow{2}[4]{*}{\textbf{Model}} & \multicolumn{4}{c}{\textbf{NeurIPS}} & & \multicolumn{4}{c}{\textbf{DBLP V10}} & & \multicolumn{4}{c}{\textbf{AI-DM}} & & \multicolumn{4}{c}{\textbf{arXiv}} & & \multicolumn{4}{c}{\textbf{PubMed}} \\
        \cmidrule{2-5}\cmidrule{7-10}\cmidrule{12-15}\cmidrule{17-20}\cmidrule{22-25}     & \multicolumn{1}{c}{TC($\uparrow$)} & \multicolumn{1}{c}{TD($\uparrow$)} & \multicolumn{1}{c}{CHI($\uparrow$)} &  \multicolumn{1}{c}{DBI($\downarrow$)} & & \multicolumn{1}{c}{TC($\uparrow$)} & \multicolumn{1}{c}{TD($\uparrow$)} & \multicolumn{1}{c}{CHI($\uparrow$)} & \multicolumn{1}{c}{DBI($\downarrow$)} & & \multicolumn{1}{c}{TC($\uparrow$)} & \multicolumn{1}{c}{TD($\uparrow$)} & \multicolumn{1}{c}{CHI($\uparrow$)} & \multicolumn{1}{c}{DBI($\downarrow$)} & & \multicolumn{1}{c}{TC($\uparrow$)} & \multicolumn{1}{c}{TD($\uparrow$)} & \multicolumn{1}{c}{CHI($\uparrow$)} & \multicolumn{1}{c}{DBI($\downarrow$)} & & \multicolumn{1}{c}{TC($\uparrow$)} & \multicolumn{1}{c}{TD($\uparrow$)} & \multicolumn{1}{c}{CHI($\uparrow$)} & \multicolumn{1}{c}{DBI($\downarrow$)}\\
        \midrule
        LDA & $\prescript{\ddag}{}{0.293}$ & $\prescript{\ddag}{}{0.118}$ & $\prescript{\ddag}{}{6.195}$ & $\prescript{\ddag}{}{6.316}$ & & $\prescript{\ddag}{}{0.409}$ & $\prescript{\ddag}{}{0.575}$ & $\prescript{\ddag}{}{68.290}$ & $\prescript{\ddag}{}{8.841}$ & & $\prescript{\ddag}{}{0.322}$ & $\prescript{\ddag}{}{0.491}$ & $\prescript{\ddag}{}{44.578}$ & $\prescript{\ddag}{}{8.931}$ & & $\prescript{\ddag}{}{0.422}$ & $\prescript{\ddag}{}{0.598}$ & $\prescript{\ddag}{}{149.606}$ & $\prescript{\ddag}{}{9.781}$ & & $\prescript{\ddag}{}{0.448}$ & $\prescript{\ddag}{}{0.581}$ & $\prescript{\ddag}{}{112.511}$ & $\prescript{\ddag}{}{7.163}$ \\
        NMF  & $\prescript{\ddag}{}{0.357}$ & $\prescript{\ddag}{}{0.071}$ & $\prescript{\ddag}{}{5.315}$ & $\prescript{\ddag}{}{7.241}$ & & $\prescript{\ddag}{}{0.380}$ & $\prescript{\ddag}{}{0.129}$ & $\prescript{\ddag}{}{55.404}$ & $\prescript{\ddag}{}{11.392}$ & & $\prescript{\ddag}{}{0.354}$ & $\prescript{\ddag}{}{0.139}$ & $\prescript{\ddag}{}{45.340}$ & $\prescript{\ddag}{}{10.485}$ & & $\prescript{\ddag}{}{0.242}$ & $\prescript{\ddag}{}{0.520}$ & $\prescript{\ddag}{}{105.767}$ & $\prescript{\ddag}{}{12.160}$ & & $\prescript{\ddag}{}{0.503}$ & $\prescript{\ddag}{}{0.245}$ & $\prescript{\ddag}{}{78.857}$ & $\prescript{\ddag}{}{10.508}$ \\
        ProdLDA  & $\prescript{\ddag}{}{0.433}$ & $\prescript{\ddag}{}{0.297}$ & $\prescript{\ddag}{}{6.768}$ & $\prescript{\ddag}{}{6.902}$ & & $\prescript{\ddag}{}{0.408}$ & $\prescript{\ddag}{}{0.186}$ & \cellcolor[rgb]{ .886,  .937,  .855}$\prescript{}{}{\textbf{1469.821}}$ & $\prescript{\ddag}{}{8.693}$ & & $\prescript{\ddag}{}{0.374}$ & $\prescript{\ddag}{}{0.614}$ & $\prescript{\ddag}{}{68.045}$ & $\prescript{\ddag}{}{8.314}$ & & $\prescript{\ddag}{}{0.466}$ & $\prescript{\ddag}{}{0.604}$ & $\prescript{\ddag}{}{15.308}$ & $\prescript{\ddag}{}{10.830}$ & & $\prescript{\ddag}{}{0.463}$ & $\prescript{\ddag}{}{0.724}$ & $\prescript{\ddag}{}{24.120}$ & $\prescript{\ddag}{}{7.991}$ \\
        DecTM & $\prescript{\ddag}{}{0.401}$ & $\prescript{\ddag}{}{0.531}$ & $\prescript{\ddag}{}{6.986}$ & $\prescript{\ddag}{}{6.868}$ & &  $\prescript{\ddag}{}{0.420}$ & $\prescript{\ddag}{}{0.598}$ & $\prescript{\ddag}{}{105.120}$ & $\prescript{\ddag}{}{8.108}$ & & $\prescript{\ddag}{}{0.363}$ & $\prescript{\ddag}{}{0.842}$ & $\prescript{\ddag}{}{77.095}$ & $\prescript{\ddag}{}{8.084}$ & & $\prescript{\ddag}{}{0.378}$ & $\prescript{\ddag}{}{0.937}$ & $\prescript{\ddag}{}{16.692}$ & $\prescript{\ddag}{}{10.389}$ & & $\prescript{\ddag}{}{0.411}$ & $\prescript{\ddag}{}{0.945}$ & $\prescript{\ddag}{}{24.469}$ & $\prescript{\ddag}{}{8.680}$ \\
        ETM & $\prescript{\ddag}{}{0.260}$ & \underline{$\prescript{\ddag}{}{0.795}$} & $\prescript{\ddag}{}{2.562}$ & $\prescript{\ddag}{}{9.147}$ & & $\prescript{\ddag}{}{0.426}$ & $\prescript{\ddag}{}{0.833}$ & $\prescript{\ddag}{}{16.662}$ & $\prescript{\ddag}{}{16.619}$ & & $\prescript{\ddag}{}{0.148}$ & $\prescript{\ddag}{}{0.836}$ & $\prescript{\ddag}{}{15.254}$ & $\prescript{\ddag}{}{15.071}$ & & $\prescript{\ddag}{}{0.322}$ & $\prescript{\ddag}{}{0.945}$ & $\prescript{\ddag}{}{2.956}$ & $\prescript{\ddag}{}{12.276}$ & & $\prescript{\ddag}{}{0.248}$ & \underline{$\prescript{\ddag}{}{0.947}$} & $\prescript{\ddag}{}{3.006}$ & $\prescript{\ddag}{}{6.763}$ \\
        NSTM & $\prescript{\ddag}{}{0.251}$ & $\prescript{\ddag}{}{0.006}$ & \underline{$\prescript{\ddag}{}{9.597}$} & $\prescript{\ddag}{}{8.626}$ & & $\prescript{\ddag}{}{0.262}$ & $\prescript{\ddag}{}{0.059}$ & $\prescript{\ddag}{}{42.418}$ & $\prescript{\ddag}{}{18.622}$ & & $\prescript{\ddag}{}{0.273}$ & $\prescript{\ddag}{}{0.091}$ & $\prescript{\ddag}{}{60.792}$ & $\prescript{\ddag}{}{14.918}$ & & $\prescript{\ddag}{}{0.392}$ & $\prescript{\ddag}{}{0.122}$ & $\prescript{\ddag}{}{15.647}$ & $\prescript{\ddag}{}{9.616}$ & & $\prescript{\ddag}{}{0.369}$ & $\prescript{\ddag}{}{0.173}$ & $\prescript{\ddag}{}{16.796}$ & \underline{$\prescript{\ddag}{}{5.425}$} \\
        TSCTM & $\prescript{\ddag}{}{0.443}$ & $\prescript{\ddag}{}{0.740}$ & $\prescript{\ddag}{}{7.814}$ & $\prescript{\ddag}{}{6.142}$ & & $\prescript{\ddag}{}{0.484}$ & $\prescript{\ddag}{}{0.634}$ & $\prescript{\ddag}{}{111.099}$ & \underline{$\prescript{\ddag}{}{7.310}$} & & $\prescript{\ddag}{}{0.422}$ & $\prescript{\ddag}{}{0.874}$ & \underline{$\prescript{\ddag}{}{85.875}$} & \underline{$\prescript{\ddag}{}{6.785}$} & & $\prescript{\ddag}{}{0.313}$ & \underline{$\prescript{\ddag}{}{0.990}$} & $\prescript{\ddag}{}{17.709}$ & \underline{$\prescript{\ddag}{}{8.049}$} & & $\prescript{\ddag}{}{0.361}$ & $\prescript{\ddag}{}{0.940}$ & $\prescript{\ddag}{}{25.149}$ & $\prescript{\ddag}{}{7.065}$ \\
        ECRTM & $\prescript{\ddag}{}{0.456}$ & $\prescript{\ddag}{}{0.554}$ & $\prescript{\ddag}{}{7.475}$ & $\prescript{\ddag}{}{6.525}$ & & \underline{$\prescript{\ddag}{}{0.559}$} & $\prescript{\ddag}{}{0.838}$ & $\prescript{\ddag}{}{82.953}$ & $\prescript{\ddag}{}{8.108}$ & & $\prescript{\ddag}{}{0.511}$ & \cellcolor[rgb]{ .886,  .937,  .855}$\prescript{}{}{\textbf{0.993}}$ & $\prescript{\ddag}{}{61.164}$ & $\prescript{\ddag}{}{8.101}$ & & $\prescript{\ddag}{}{0.443}$ & $\prescript{\ddag}{}{0.853}$ & $\prescript{\ddag}{}{24.801}$ & $\prescript{\ddag}{}{8.332}$ & & $\prescript{\ddag}{}{0.410}$ & $\prescript{\ddag}{}{0.910}$ & $\prescript{\ddag}{}{23.530}$ & $\prescript{\ddag}{}{7.586}$ \\
        BERTopic & $\prescript{\ddag}{}{0.454}$ & $\prescript{\ddag}{}{0.205}$ & $\prescript{\ddag}{}{8.288}$ & \underline{$\prescript{\ddag}{}{5.674}$} & & $\prescript{\ddag}{}{0.440}$ & $\prescript{\ddag}{}{0.301}$ & $\prescript{\ddag}{}{84.408}$ & $\prescript{\ddag}{}{7.829}$ & & $\prescript{\ddag}{}{0.378}$ & $\prescript{\ddag}{}{0.362}$ & $\prescript{\ddag}{}{64.164}$ & $\prescript{\ddag}{}{7.335}$ & & \underline{$\prescript{\ddag}{}{0.608}$} & $\prescript{\ddag}{}{0.428}$ & $\prescript{\ddag}{}{148.405}$ & $\prescript{\ddag}{}{9.038}$ & & $\prescript{\ddag}{}{0.555}$ & $\prescript{\ddag}{}{0.414}$ & $\prescript{\ddag}{}{115.203}$ & $\prescript{\ddag}{}{8.009}$ \\
        FASTopic & \underline{$\prescript{\ddag}{}{0.527}$} & $\prescript{\ddag}{}{0.753}$ & $\prescript{\ddag}{}{7.983}$ & $\prescript{\ddag}{}{6.636}$ & & $\prescript{\ddag}{}{0.551}$ & \underline{$\prescript{\ddag}{}{0.850}$} & $\prescript{\ddag}{}{105.247}$ & $\prescript{\ddag}{}{8.184}$ & & \underline{$\prescript{\ddag}{}{0.560}$} & $\prescript{\ddag}{}{0.967}$ & $\prescript{\ddag}{}{81.451}$ & $\prescript{\ddag}{}{8.307}$ & & $\prescript{\ddag}{}{0.564}$ & $\prescript{\ddag}{}{0.367}$ & \underline{$\prescript{\ddag}{}{253.636}$} & $\prescript{\ddag}{}{9.206}$ & & \underline{$\prescript{\ddag}{}{0.578}$} & $\prescript{\ddag}{}{0.350}$ & \underline{$\prescript{\ddag}{}{223.432}$} & $\prescript{\ddag}{}{7.564}$ \\
        \midrule
        \textbf{\methodname} & \cellcolor[rgb]{ .886,  .937,  .855}\textbf{0.657} & \cellcolor[rgb]{ .886,  .937,  .855}\textbf{0.973} & \cellcolor[rgb]{ .886,  .937,  .855}\textbf{11.049} & \cellcolor[rgb]{ .886,  .937,  .855}\textbf{5.304} & & \cellcolor[rgb]{ .886,  .937,  .855}\textbf{0.753} & \cellcolor[rgb]{ .886,  .937,  .855}\textbf{0.988} & \underline{264.599} & \cellcolor[rgb]{ .886,  .937,  .855}\textbf{4.843} & & \cellcolor[rgb]{ .886,  .937,  .855}\textbf{0.648} & \underline{0.991} & \cellcolor[rgb]{ .886,  .937,  .855}\textbf{157.785} & \cellcolor[rgb]{ .886,  .937,  .855}\textbf{5.369} & & \cellcolor[rgb]{ .886,  .937,  .855}\textbf{0.779} & \cellcolor[rgb]{ .886,  .937,  .855}\textbf{0.993} & \cellcolor[rgb]{ .886,  .937,  .855}\textbf{342.564} & \cellcolor[rgb]{ .886,  .937,  .855}\textbf{5.984}  & & \cellcolor[rgb]{ .886,  .937,  .855}\textbf{0.725} & \cellcolor[rgb]{ .886,  .937,  .855}\textbf{0.964} & \cellcolor[rgb]{ .886,  .937,  .855}\textbf{7820.389} & \cellcolor[rgb]{ .886,  .937,  .855}\textbf{3.086} \\
        \bottomrule
        
    \end{tabular}
    }
    \caption{
        Topic quality results on different datasets (with topic numbers \(K=100\)). 
        The superscript $\ddag$ means the gains of SciTopic are statistically significant at 0.05 level.
    }
    \label{tab_topic_quality}%
\end{table*}%

\begin{table}[!t]
    \centering
    \setlength{\tabcolsep}{1.5mm}
    \resizebox{\linewidth}{!}
    {
    \begin{tabular}{lrrrrrrrrrrrrrrr}
        \toprule
        \multirow{2}[4]{*}{\textbf{Model}} & \multicolumn{3}{c}{\textbf{AG News}} & & \multicolumn{3}{c}{\textbf{20 News Groups}}\\
        \cmidrule{2-4}\cmidrule{6-8} & \multicolumn{1}{c}{ACC($\uparrow$)} & \multicolumn{1}{c}{NMI($\uparrow$)} & \multicolumn{1}{c}{ARI($\uparrow$)} & & \multicolumn{1}{c}{ACC($\uparrow$)} & \multicolumn{1}{c}{NMI($\uparrow$)} & \multicolumn{1}{c}{ARI($\uparrow$)}\\
        \midrule
        LDA & $74.05 \text{\tiny $\pm$ 8.51}$ & $47.17 \text{\tiny $\pm$ 9.32}$ & $49.01 \text{\tiny $\pm$ 10.49}$ & & $29.05 \text{\tiny $\pm$ 0.85}$ & $31.63 \text{\tiny $\pm$ 1.22}$ & $13.34 \text{\tiny $\pm$ 2.70}$ \\
        NMF & $34.05 \text{\tiny $\pm$ 2.48}$ & $4.59 \text{\tiny $\pm$ 1.01}$ & $2.13 \text{\tiny $\pm$ 1.08}$ & & $12.42 \text{\tiny $\pm$ 1.91}$ & $12.86 \text{\tiny $\pm$ 2.72}$ & $0.48 \text{\tiny $\pm$ 0.32}$ \\
        ProdLDA & $80.93 \text{\tiny $\pm$ 0.04}$ & $56.51 \text{\tiny $\pm$ 0.09}$ & $60.91 \text{\tiny $\pm$ 0.08}$ & & $37.42 \text{\tiny $\pm$ 3.83}$ & $45.67 \text{\tiny $\pm$ 3.35}$ & $23.89 \text{\tiny $\pm$ 3.09}$ \\
        DecTM & $55.63 \text{\tiny $\pm$ 2.11}$ & $40.04 \text{\tiny $\pm$ 1.88}$ & $36.17 \text{\tiny $\pm$ 2.65}$ & & $36.57 \text{\tiny $\pm$ 0.55}$ & $46.18 \text{\tiny $\pm$ 0.44}$ & $22.90 \text{\tiny $\pm$ 0.85}$ \\
        ETM & $26.14 \text{\tiny $\pm$ 0.00}$ & $0.00 \text{\tiny $\pm$ 0.00}$ & $0.00 \text{\tiny $\pm$ 0.00}$ & & $5.35 \text{\tiny $\pm$ 0.01}$ & $0.10 \text{\tiny $\pm$ 0.01}$ & $0.00 \text{\tiny $\pm$ 0.00}$ \\
        NSTM & $26.14 \text{\tiny $\pm$ 0.00}$ & $0.01 \text{\tiny $\pm$ 0.01}$ & $0.00 \text{\tiny $\pm$ 0.00}$ & & $16.92 \text{\tiny $\pm$ 6.76}$ & $17.02 \text{\tiny $\pm$ 6.61}$ & $2.34 \text{\tiny $\pm$ 2.98}$ \\
        TSCTM & $79.63 \text{\tiny $\pm$ 1.22}$ & $53.91 \text{\tiny $\pm$ 1.42}$ & $55.89 \text{\tiny $\pm$ 1.50}$ & & $40.60 \text{\tiny $\pm$ 2.22}$ & $44.06 \text{\tiny $\pm$ 1.24}$ & $15.71 \text{\tiny $\pm$ 0.63}$ \\
        ECRTM & $78.69 \text{\tiny $\pm$ 2.44}$ & $54.05 \text{\tiny $\pm$ 2.63}$ & $54.88 \text{\tiny $\pm$ 4.02}$ & & $25.70 \text{\tiny $\pm$ 2.29}$ & $31.00 \text{\tiny $\pm$ 0.69}$ & $12.26 \text{\tiny $\pm$ 0.21}$ \\
        Bertopic & $35.93 \text{\tiny $\pm$ 8.62}$ & $12.88 \text{\tiny $\pm$ 10.55}$ & $7.03 \text{\tiny $\pm$ 6.07}$ & & $29.78 \text{\tiny $\pm$ 1.98}$ & $28.57 \text{\tiny $\pm$ 1.60}$ & $11.58 \text{\tiny $\pm$ 5.66}$ \\
        FASTopic & \underline{$83.48 \text{\tiny $\pm$ 00.08}$} & \underline{$59.10 \text{\tiny $\pm$ 00.10}$} & \underline{$62.48 \text{\tiny $\pm$ 00.15}$} & & \underline{$51.65 \text{\tiny $\pm$ 0.97}$} & \underline{$56.32 \text{\tiny $\pm$ 1.13}$} & \underline{$39.49 \text{\tiny $\pm$ 1.84}$} \\
        \midrule
        \textbf{\methodname} & \cellcolor[rgb]{ .886,  .937,  .855}\textbf{$\textbf{85.29} \text{\tiny $\pm$ 00.01}$} & \cellcolor[rgb]{ .886,  .937,  .855}\textbf{$\textbf{61.96} \text{\tiny $\pm$ 00.01}$} & \cellcolor[rgb]{ .886,  .937,  .855}\textbf{$\textbf{65.94} \text{\tiny $\pm$ 00.01}$} & & \cellcolor[rgb]{ .886,  .937,  .855}\textbf{$\textbf{70.88} \text{\tiny $\pm$ 0.60}$} & \cellcolor[rgb]{ .886,  .937,  .855}\textbf{$\textbf{68.32} \text{\tiny $\pm$ 0.46}$} & \cellcolor[rgb]{ .886,  .937,  .855}\textbf{$\textbf{55.71} \text{\tiny $\pm$ 0.74}$}\\
        \bottomrule
        
    \end{tabular}
    }
    \caption{
        Clustering performance on labeled datasets: AG News and 20 News Groups.
    }
    \vspace{-5mm}
    \label{tab:labeled_clustering_case_study}%
    
\end{table}%

\begin{figure*}[htbp]
\centering
\includegraphics[width=1.0\linewidth]{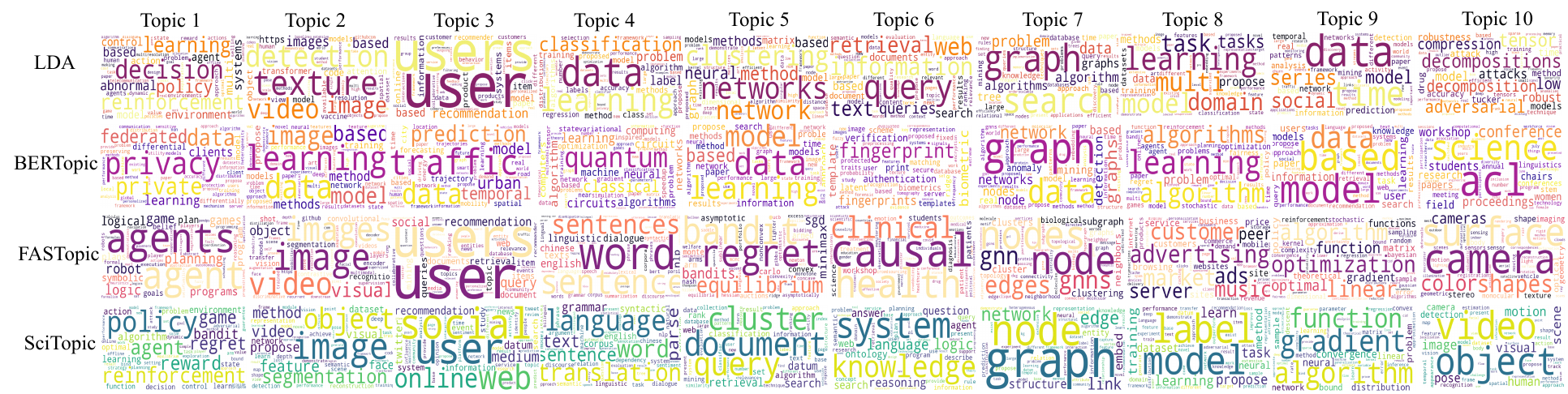}
\caption{WordCloud visualization on AI-DM of top 50 words per topic across LDA, BERTopic, Fastopic, and SciTopic, when topic number equals 10.}
\label{fig:topic-analysis}
\end{figure*}

\subsection{Experiment Setup}
\noindent\textbf{Baseline Methods.}
We evaluate our model against ten benchmark models encompassing a spectrum of traditional and advanced neural topic modeling techniques:
\textbf{(\romannumeral 1) LDA}~\cite{blei2003latent}, a classical probabilistic generative model for topic discovery.  
\textbf{(\romannumeral 2) NMF}~\cite{lee2000algorithms}, a matrix decomposition technique used for topic discovery.  
\textbf{(\romannumeral 3) ProdLDA}~\cite{srivastava2017autoencoding}, a neural topic model based on variational autoencoders, incorporating product-of-experts priors for topic generation.  
\textbf{(\romannumeral 4) DecTM}~\cite{wu2021discovering} decouples topic modeling into separate modules for word and document distributions.  
\textbf{(\romannumeral 5) ETM}~\cite{dieng2020topic} combines word embeddings with generative topic modeling.  
\textbf{(\romannumeral 6) NSTM}~\cite{zhao2020neural} utilizes optimal transport theory to mitigate semantic bias in neural topic models.  
\textbf{(\romannumeral 7) TSCTM}~\cite{wu2022mitigating} employs contrastive learning for short text topic modeling.  
\textbf{(\romannumeral 8) ECRTM}~\cite{wu2023effective} prevents topic collapse through embedding clustering regularization.  
\textbf{(\romannumeral 9) BERTopic}~\cite{grootendorst2022bertopic} leverages pre-trained transformer-based embeddings for topic generation.
\textbf{(\romannumeral 10) FASTopic}~\cite{wu2024fastopic} introduces dual semantic-relation reconstruction for adaptive, stable, and transferable topic discovery. 
We fine-tune the hyperparameters of these baselines under different datasets and topic numbers.

\noindent\textbf{Implementation Details.} 
We use the BGE-M3 model for embedding generation, fine-tuned on domain-specific data to optimize semantic representation \cite{chen2024bge}. 
Clustering uses the K-Means algorithm \cite{macqueen1967some} and the Llama-3.1-70B model \cite{dubey2024llama}. 
Parameters are set as $\alpha=1$ and $\lambda=0.5$, and experiments are conducted on two NVIDIA A100-80GB GPUs. 

\begin{figure*}[!t]
    \centering
    \includegraphics[width=1\linewidth]{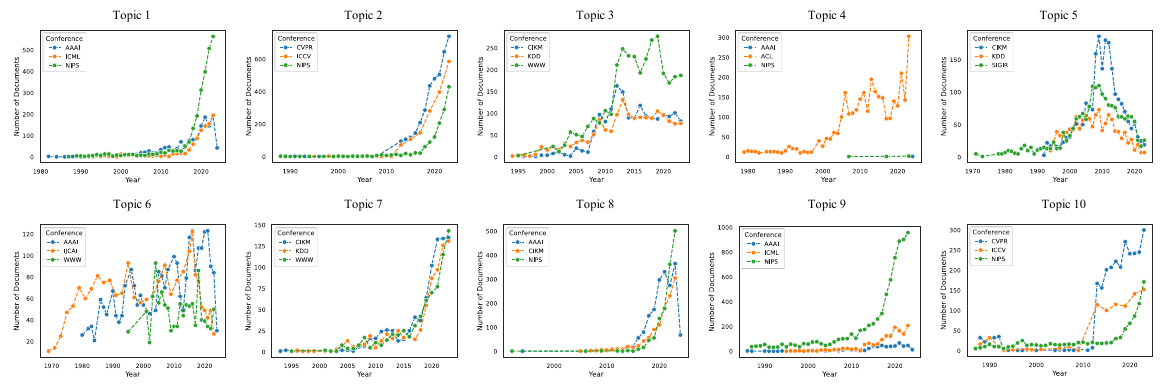}
    \caption{Temporal evolution of all topics on AI-DM.}
    \label{fig:topic_trends_all}
\end{figure*}

\noindent\textbf{Evaluation Metrics.}
In this section, we describe the evaluation metrics used to assess the performance of the proposed method. The metrics are categorized into topic discovery evaluation metrics and clustering evaluation metrics.

\noindent\textbf{\textit{Topic Discovery Evaluation Metrics.}} 
\textbf{(\romannumeral 1) Topic Coherence (TC).}
Topic coherence measures the semantic similarity of the most significant words within each topic. Specifically, we use the \(\text{C}_\text{V}\) metric. The formula for \(\text{C}_\text{V}\) is:
\begin{equation}
    \text{C}_\text{V} = \frac{1}{|W|} \sum_{w_i \in W} \sum_{w_j \in W, j > i} \text{NPMI}(w_i, w_j) \cdot \log(P(w_i, w_j)),
\end{equation}
where \(W = \{w_1, w_2, \dots, w_T\}\) represents the set of top \(T\) words for a topic, \(P(w_i, w_j)\) is the co-occurrence probability of words \(w_i\) and \(w_j\), and \(\text{NPMI}(w_i, w_j)\) is the normalized pointwise mutual information score. 

\textbf{(\romannumeral 2) Topic Diversity (TD).}
Topic diversity reflects the uniqueness and coverage of topics by measuring the variety of words across all topics. The TD is defined as:
\begin{equation}
    \text{TD} = \frac{\text{Number of unique words across all topics}}{k \times \text{Number of topics}},
\end{equation}
where \(k\) is the number of top words considered for each topic. A higher \(\text{TD}\) value indicates greater topic diversity and significantly less overlap among the topics.

\noindent\textbf{\textit{Clustering Evaluation Metrics.}} 
\textbf{(\romannumeral 3) Calinski-Harabasz Index (CHI).}
The Calinski-Harabasz Index evaluates the ratio of the between-cluster dispersion to the within-cluster dispersion, providing a widely used statistical criterion for assessing clustering quality. The CHI is computed as:
\begin{equation}
    \text{CHI} = \frac{\text{trace}(B_k)}{\text{trace}(W_k)} \cdot \frac{n-k}{k-1},
\end{equation}
where \(B_k\) and \(W_k\) are the between-cluster and within-cluster scatter matrices, respectively, \(n\) is the number of samples, and \(k\) is the number of clusters. Higher \(\text{CHI}\) values indicate better clustering quality.

\textbf{(\romannumeral 4) Davies-Bouldin Index (DBI).}
The Davies-Bouldin Index evaluates the ratio of intra-cluster distances to inter-cluster distances. The DBI is defined as:
\begin{equation}
    \text{DBI} = \frac{1}{N} \sum_{i=1}^N \max_{j \neq i} \left( \frac{\sigma_i + \sigma_j}{d_{ij}} \right),
\end{equation}
where \(N\) is the number of clusters, \(\sigma_i\) and \(\sigma_j\) represent the intra-cluster distances for clusters \(i\) and \(j\), respectively, and \(d_{ij}\) is the distance between the centroids of clusters \(i\) and \(j\). Lower \(\text{DBI}\) values indicate better clustering performance.

\subsection{Overall Performance}
\textbf{Quantitative Analysis.} \textcolor{black}{Table~\ref{tab_topic_quality} compares SciTopic with ten baselines on five datasets for $K=100$. SciTopic consistently outperforms traditional methods (e.g., LDA, NMF) and neural topic models (e.g., BERTopic, FASTopic) by achieving higher TC and TD. Additionally, SciTopic demonstrates a better balance between coherence and diversity compared to advanced models like ECRTM and TSCTM, which often show variability in one of these metrics. On average, SciTopic improves TC by \textbf{21.8\%}, TD by \textbf{14.6\%}, and CHI by \textbf{5.61\%} over the second-best method, highlighting its superior ability to handle complex thematic structures.}


\begin{table}[!t]
    \centering
    \resizebox{\linewidth}{!}{
    
        \begin{tabular}{llc}
            \toprule
             Topic ID & Top-3 keywords & Top-3 source venues \\
            \midrule
            \multirow{1}{*}{Topic 1} & policy, reinforcement, agent & AAAI, ICML, NIPS \\
            \multirow{1}{*}{Topic 2} & image, object, segmentation & CVPR, ICCV, NIPS \\
            \multirow{1}{*}{Topic 3} & user, social, web & CIKM, KDD, WWW \\
            \multirow{1}{*}{Topic 4} & language, translation, word & AAAI, ACL, NIPS \\
            \multirow{1}{*}{Topic 5} & document, query, cluster & CIKM, KDD, SIGIR \\
            \multirow{1}{*}{Topic 6} & system, knowledge, logic & AAAI, IJCAI, WWW \\
            \multirow{1}{*}{Topic 7} & graph, node, network & CIKM, KDD, WWW \\
            \multirow{1}{*}{Topic 8} & model, label, learning & AAAI, CIKM, NIPS \\
            \multirow{1}{*}{Topic 9} & gradient, algorithm, function & AAAI, ICML, NIPS \\
            \multirow{1}{*}{Topic 10} & object, video, motion & CVPR, ICCV, NIPS \\
            
            \bottomrule
        \end{tabular}
    }
    \caption{Keywords and venue.}
    \vspace{-5mm}
    \label{tab:keyword_venue}
\end{table}
\noindent\textbf{Semantic Analysis.} 
We qualitatively compared LDA, BERTopic, Fastopic, and SciTopic across topic clarity, distinctiveness, and keyword diversity. As shown in Figure \ref{fig:topic-analysis}, SciTopic outperforms others with precise, semantically focused topics, such as distinct clusters for reinforcement learning (reinforcement, policy, agent) and graph neural networks (graph, node, network).
It achieves superior topic separation with minimal overlap while balancing diversity by capturing broad, domain-specific terms with remarkable consistency.
BERTopic performs well in clarity and diversity, especially in fields like quantum computing, but occasionally exhibits subtle keyword overlap.
Fastopic covers diverse concepts but includes low-relevance terms that dilute focus, while LDA suffers from generic, overlapping terms with low specificity.


\subsection{Case Study}
\label{sec:case}
We present three case studies to thoroughly assess the effectiveness and interpretability of \textbf{\methodname}: the first focuses on clustering accuracy using ground-truth labels, the second examines source venues of papers associated with each topic, and the third investigates topics' temporal evolution over time.

\noindent\textbf{Ground-truth Clustering Validation.} 
As shown in Table~\ref{tab:labeled_clustering_case_study}, we evaluate clustering quality on two labeled corpora, AG News and 20 News Groups, using Accuracy (ACC), Normalized Mutual Information (NMI), and Adjusted Rand Index (ARI). 
\textbf{SciTopic} consistently outperforms baseline models across diverse datasets, especially on the more challenging 20 News Groups benchmark, clearly demonstrating its strong ability to form coherent and label-aligned topic clusters.

\noindent\textbf{Topic-venue Consistency.}
We compare the extracted keywords of each topic with the top venues of papers from that topic, as shown in Table~\ref{tab:keyword_venue}. For instance, \textit{Topic 2} features image, object, and segmentation, and most papers on this topic are published on CVPR and ICCV, which focus on computer vision; \textit{Topic 3} features user, social, and web, and most papers are from CIKM, KDD, and WWW, which focus on graph and web data mining.  

\noindent\textbf{Topic Evolution Dynamics.}
\begin{figure}[t!]
    \centering
    \subfloat[\(TC(\uparrow)\)]{
        \includegraphics[width=0.42\linewidth]{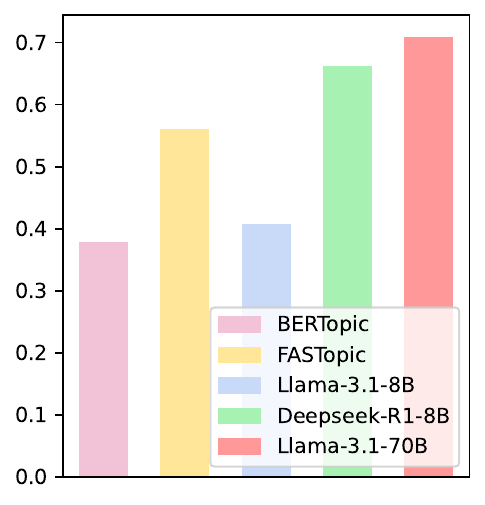}
        \label{fig:sub1_tc}
    }
    \subfloat[\(TD(\uparrow)\)]{
        \includegraphics[width=0.42\linewidth]{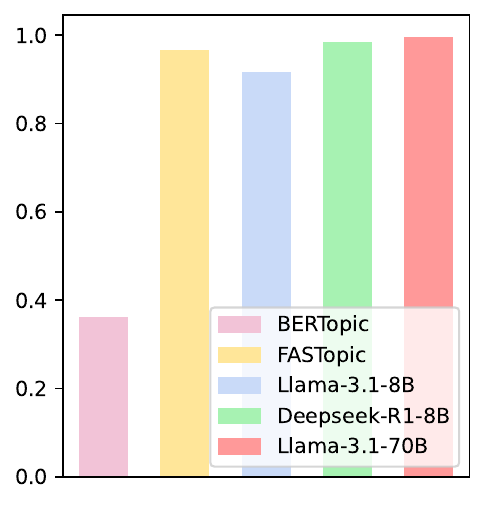}
        \label{fig:sub2_td}
    }
    \\
    \subfloat[\(CHI(\uparrow)\)]{
        \includegraphics[width=0.42\linewidth]{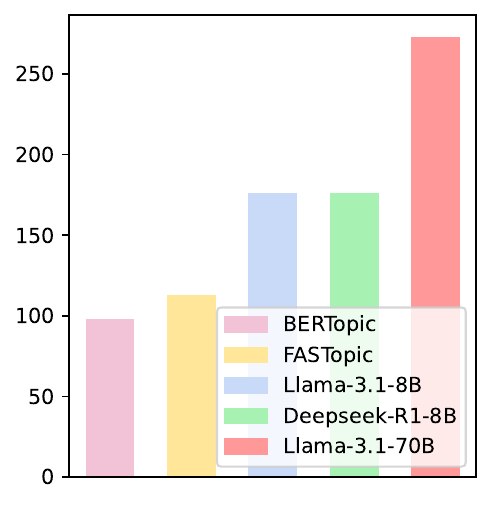}
        \label{fig:sub3_chi}
    }
    \subfloat[\(DBI(\downarrow)\)]{
        \includegraphics[width=0.4\linewidth]{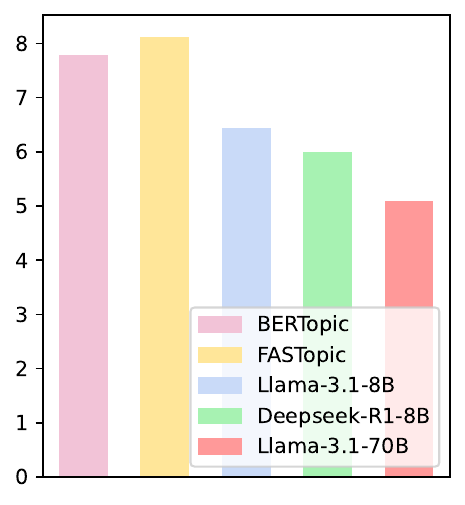}
        \label{fig:sub4_dbi}
    }
    \caption{Parameter sensitivity analysis on Topic Coherence and Topic Diversity.}
    \vspace{-5mm}
    \label{fig:model_comp}
\end{figure}
To assess topic interpretability over time, we analyze the dynamics of \textit{Topic 2} (computer vision) and \textit{Topic 5} (information retrieval), as shown in Figure~\ref{fig:topic_trends_all}. 
\textit{Topic 2} shows marked growth after 2013, which reflects the rapid advancement of CNNs in the computer vision field after the breakthrough of AlexNet in 2012. 
And \textit{Topic 5} shows a clear peak around 2008-2012, followed by a gradual decline. 
This trend was likely associated with peak research interest in information retrieval during the rapid expansion of search engines such as Google, Bing and Baidu. 

\subsection{Ablation Study}
\textbf{Framework Component Evaluation.}
To assess the effectiveness of individual components in \textbf{SciTopic}, we conducted an ablation study across both model and dataset configurations (Table~\ref{tab:ablation}). 
For the model, we denote \textbf{SciTopic w/o FT} as the variant without fine-tuning and \textbf{SciTopic w/o Entropy} as the model without sampling based on entropy. We further define \textbf{SciTopic w/o Distance} as the variant that replaces LLM-guided triplets with traditional distance-based triplet sampling for comparison purposes. Similarly, for the dataset, \textbf{SciTopic w/o Title (T)}, \textbf{SciTopic w/o Abstract (A)}, and \textbf{SciTopic w/o Metadata (M)} indicate the removal of titles, abstracts, and metadata, respectively. Variants such as \textbf{SciTopic w/o TM} (removing both title and metadata) and \textbf{SciTopic w/o AM} (removing abstract and metadata) are also evaluated. As shown in Table~\ref{tab:ablation}, every component contributes significantly to the model’s overall performance, and removing any of them leads to noticeable performance degradation, clearly highlighting their critical importance in ensuring efficiency, robustness, and reliable topic discovery effectiveness.

\noindent \textbf{Input Component Evaluation.}
As shown in Table~\ref{tab:ablation}, both \textbf{Title}, \textbf{Abstract}, and \textbf{Metadata} contribute to performance. 
Removing \textbf{Title} (TC=0.519, TD=0.872) or \textbf{Abstract} (TC=0.544, TD=0.890) caused clear drops, while excluding \textbf{Metadata} had a milder impact (TC=0.613, TD=0.951). 
More severe declines appeared when combining removals, e.g., \textbf{w/o TM} (TC=0.501, TD=0.858) and \textbf{w/o AM} (TC=0.507, TD=0.907), underscoring the complementary role of these components in enhancing topic coherence and diversity.

\noindent \textbf{Reliance Study on LLM's Parameter and Capacity.}
A core part of \textbf{SciTopic} is the LLM guided clustering, where we employ Llama-3.1-70B for the triplet task, and to investigate the methods' reliance on the LLM, we replace the 70B model with smaller variants. As shown in Figure~\ref{fig:model_comp}, when replacing the LLM with Llama-3.1-8B and Deepseek-R1-8B, we can observe decreases in all four measurements. However, even with smaller LLMs, the performance is superior to that of other methods, especially with the cutting-edge Deepseek-R1-8B model, which exhibits comparable results with the Llama-3.1-70B model. These results demonstrate the effectiveness of \textbf{SciTopic}'s methodology, suggesting that \textbf{SciTopic} benefits from but does not rely on very large LLMs and is still applicable in low-resource scenarios. Given that LLMs are rapidly evolving and smaller LLMs are getting more powerful thanks to techniques such as knowledge distillation, we could expect to adapt \textbf{SciTopic} with the latest small-sized LLMs while achieving favorable capacity.

\begin{table}[!t]
    \centering
    \resizebox{\linewidth}{!}{
    
        \begin{tabular}{llcccc}
            \toprule
             & Variants & TC ($\uparrow$) & TD ($\uparrow$) & CHI ($\uparrow$) & DBI ($\downarrow$) \\
            \midrule
            \multirow{2}{*}{Model} & 
            SciTopic w/o FT & $\prescript{\ddag}{}{0.620}$ & $\prescript{\ddag}{}{0.885}$ & $\prescript{\ddag}{}{	106.501}$ & $\prescript{\ddag}{}{6.025}$ \\
            & SciTopic w/o Entropy & $\prescript{\ddag}{}{0.625}$ & $\prescript{\ddag}{}{0.979}$ & $\prescript{\ddag}{}{	179.391}$ & $\prescript{\ddag}{}{5.095}$ \\
            & SciTopic r/p Distance & $\prescript{\ddag}{}{0.539}$ &	$\prescript{\ddag}{}{0.970}$ & \cellcolor[rgb]{ .886,  .937,  .855}{\textbf{573.937}} & \cellcolor[rgb]{ .886,  .937,  .855}{\textbf{4.474}} \\
            \cmidrule{1-6}
            \multirow{5}{*}{Dataset} & 
            SciTopic w/o Title & $\prescript{\ddag}{}{0.519}$ & $\prescript{\ddag}{}{0.872}$ & $\prescript{\ddag}{}{73.029}$ & $\prescript{\ddag}{}{5.625}$ \\
            & SciTopic w/o Abstract & $\prescript{\ddag}{}{0.544}$ & $\prescript{\ddag}{}{0.890}$ & $\prescript{\ddag}{}{69.949}$ & $\prescript{\ddag}{}{5.763}$ \\
            & SciTopic w/o Meta & $\prescript{\ddag}{}{0.613}$ & $\prescript{\ddag}{}{0.951}$ & $\prescript{\ddag}{}{83.566}$ & $\prescript{\ddag}{}{5.318}$ \\
            & SciTopic w/o TM & $\prescript{\ddag}{}{0.501}$ & $\prescript{\ddag}{}{0.858}$ & \underline{$\prescript{\ddag}{}{273.684}$} & \underline{$\prescript{\ddag}{}{5.143}$} \\
            & SciTopic w/o AM & $\prescript{\ddag}{}{0.507}$ & $\prescript{\ddag}{}{0.907}$ & $\prescript{\ddag}{}{204.722}$ & $\prescript{\ddag}{}{5.892}$ \\
            \cmidrule{1-6}
             & \textbf{\methodname} & \cellcolor[rgb]{ .886,  .937,  .855}\textbf{0.648} & \cellcolor[rgb]{ .886,  .937,  .855}\textbf{0.991} & 157.785 & 5.369 \\
            \bottomrule
        \end{tabular}
    }
    \caption{Performance of SciTopic variants via component ablation (with \(K=100\)).}
    \vspace{-5mm}
    \label{tab:ablation}
\end{table}

\subsection{Parameter Sensitivity Analysis}
\begin{figure}[t!]
    \centering
    \subfloat[\(\alpha\)]{
        \includegraphics[width=0.49\linewidth]{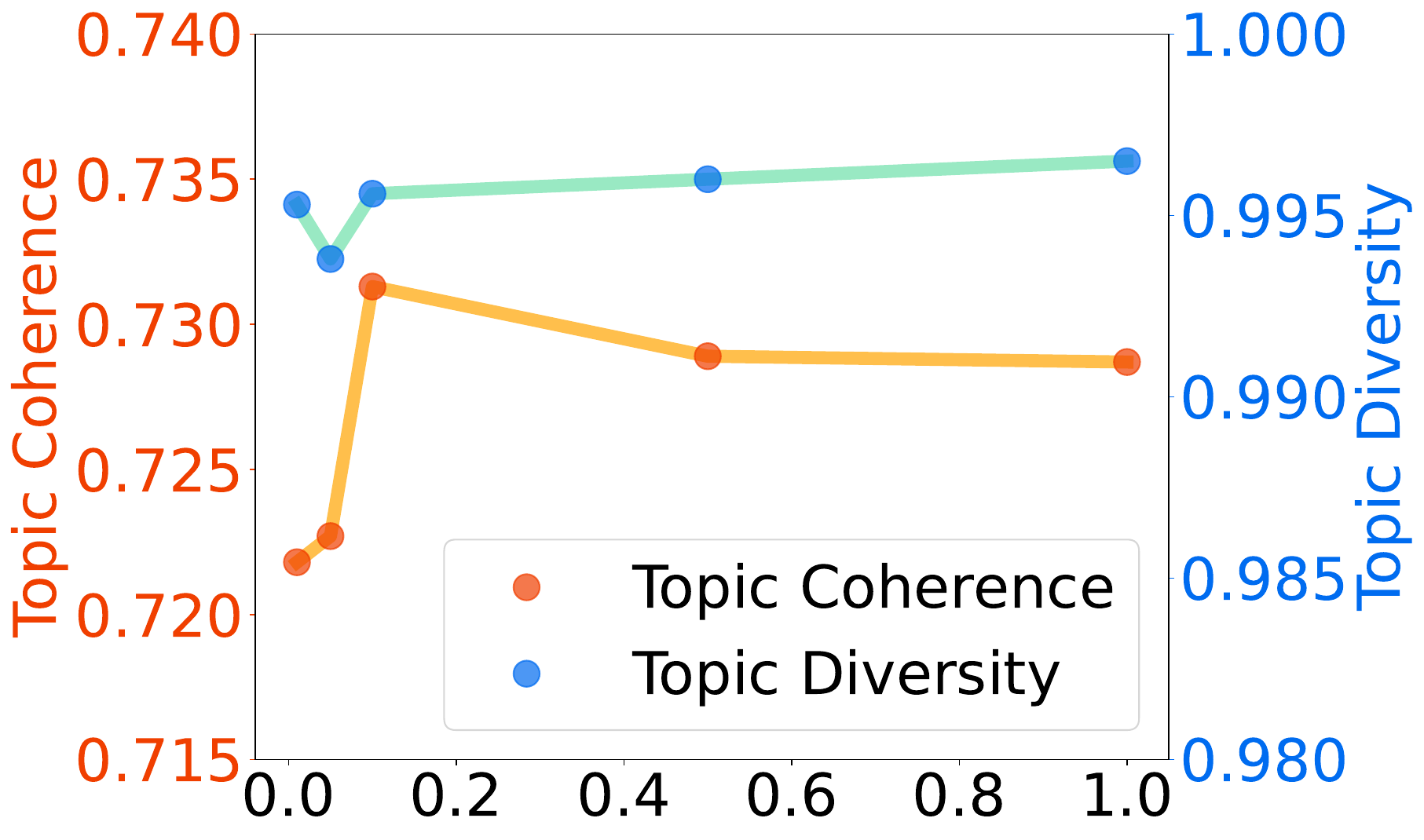}
        \label{fig:sub1_alpha}
    }
    \subfloat[\(\lambda\)]{
        \includegraphics[width=0.49\linewidth]{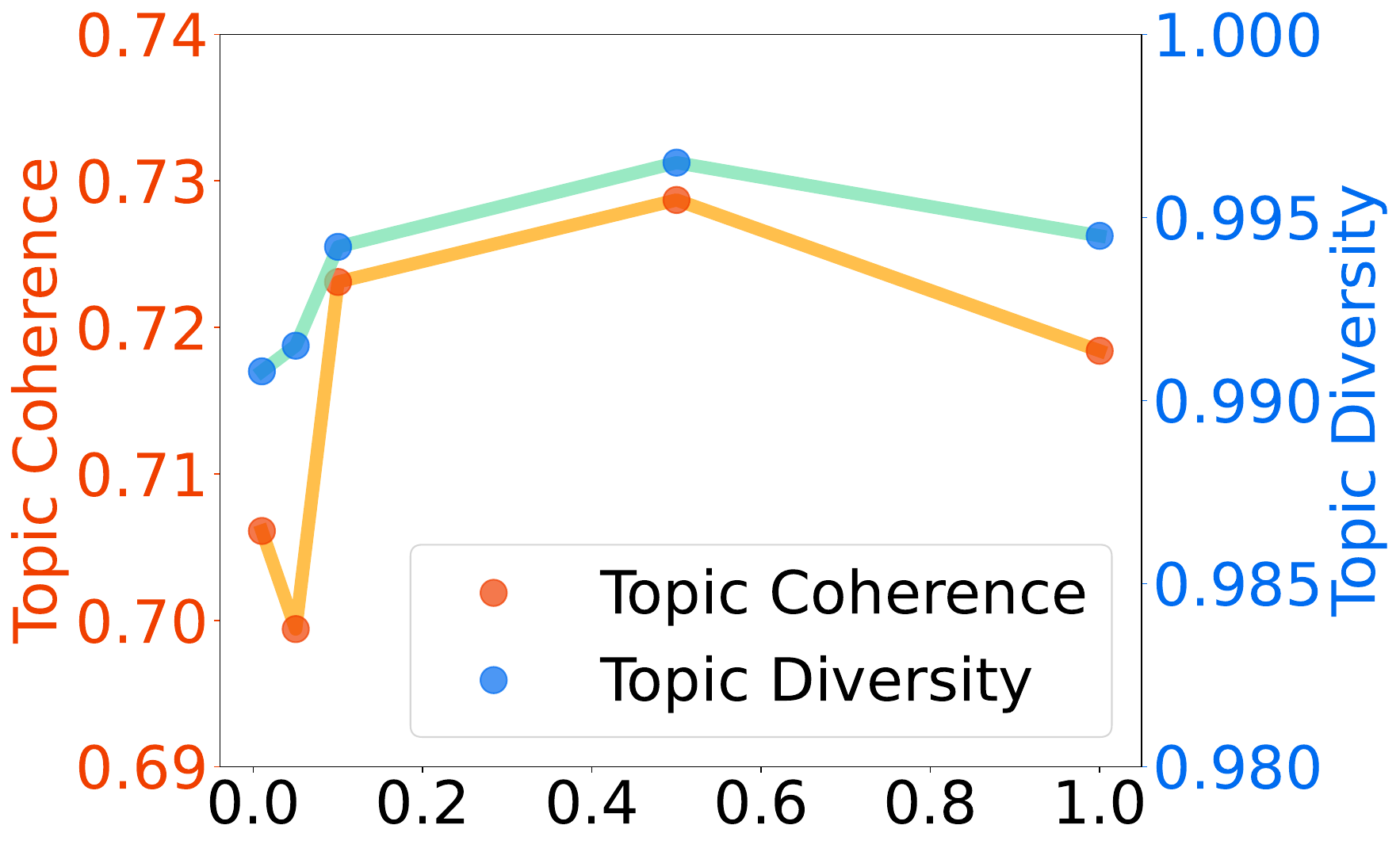}
        \label{fig:sub2_lambda}
    }
    \caption{Parameter sensitivity analysis on Topic Coherence and Topic Diversity.}
    \label{fig:sensitivity}
\end{figure}

To assess the impact of the parameters alpha and lambda on our model's efficacy, we executed a parameter sensitivity analysis focusing on TC and TD. Figure \ref{fig:sensitivity} displays the outcomes of these evaluations. The parameter alpha was varied from 1.0 down to 0.01, and lambda was similarly adjusted within the same range.
Our analysis revealed that the variation in TC and TD across different parameter settings was marginal and relatively insignificant. Notably, the highest TC value obtained was 0.7313, occurring at $\alpha$ = 0.1 and $\lambda$ = 0.5. Despite these variations, the model consistently demonstrated considerable robustness and stability, indicating a remarkably low sensitivity to even moderate changes in alpha and lambda parameters. This robustness therefore suggests highly consistent performance across a wide spectrum of parameter values, further underscoring the model's reliability, generalizability, and effectiveness in diverse real-world operational contexts.

\subsection{Computational Efficiency Analysis}
\textbf{Processing Time Across Datasets.}
To provide a fair comparison of computational cost across datasets, we counted the processing time required to analyze 20\% of each dataset using the same experimental setup. 
Table~\ref{tab:time_cost_20percent} reports the Statistical time to process 20\% of each dataset. This statistic reflects the relative scale and expected runtime of applying the \textbf{SciTopic} framework across different domains.

\begin{table}[t!]
\centering
\resizebox{\linewidth}{!}{\begin{tabular}{lrrr}
\toprule
\textbf{Dataset} & \textbf{Total Papers} & \textbf{20\% Sample} & \textbf{Time (min)} \\
\midrule
AI-DM     & 78,020     & 15,604   & 100.0  \\
DBLP V10  & 100,000    & 20,000   & 128.2  \\
NeurIPS   & 7,241      & 1,448    & 9.3    \\
arXiv     & 20,000     & 4,000    & 25.6   \\
PubMed    & 142,432    & 28,486   & 182.6  \\
\bottomrule
\end{tabular}}
\vspace{-1mm}
\caption{Processing time on 20\% of each dataset.}
\vspace{-3mm}
\label{tab:time_cost_20percent}
\end{table}

\begin{figure}[t!]
    \centering
    \includegraphics[width=0.9\linewidth]{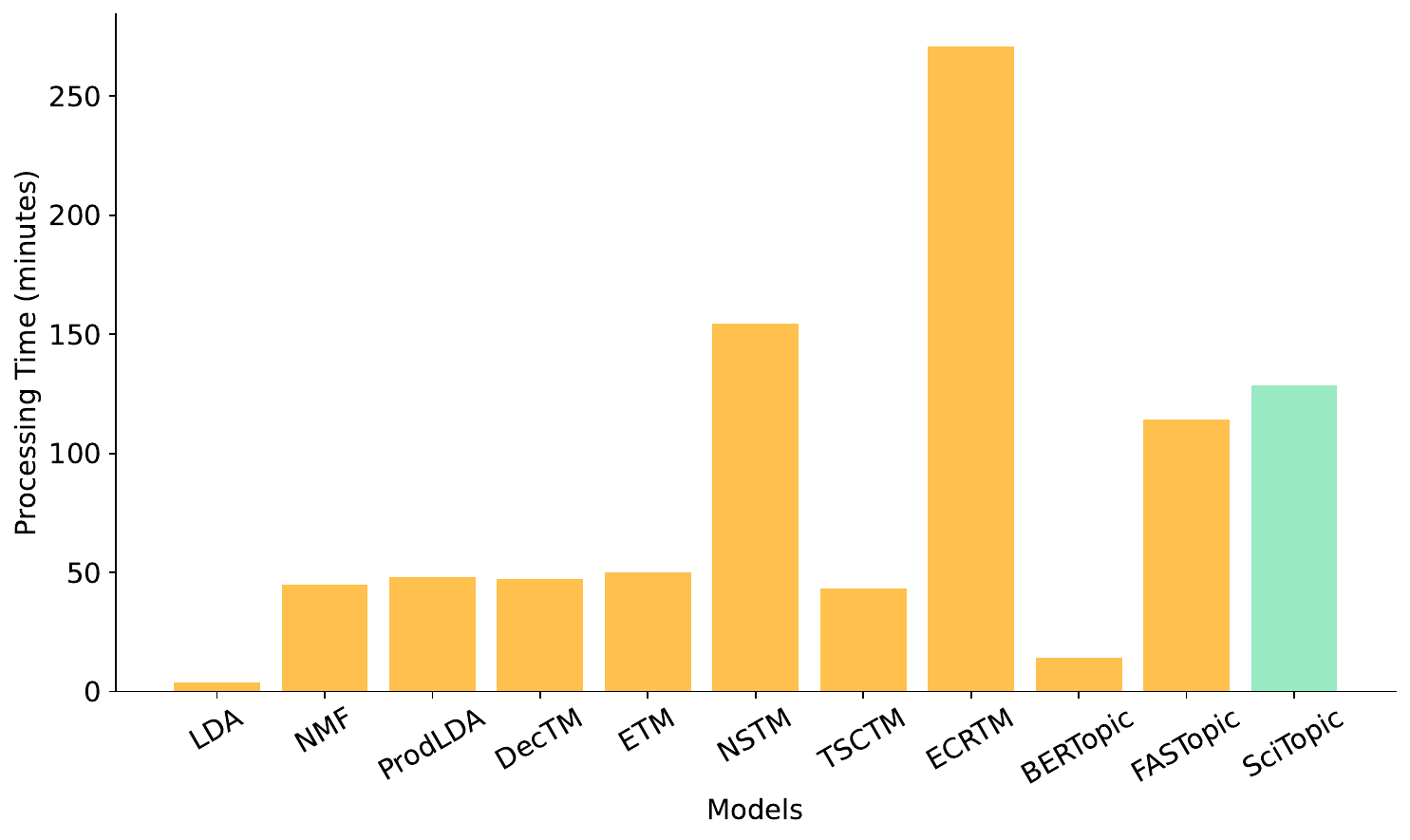}

    \vspace{-1mm}
    \caption{Runtime comparison across topic modeling methods on the AI-DM dataset (in minutes).}
    \vspace{-3mm}
    \label{fig:ai_dm_time}
\end{figure}

\textbf{Runtime Comparison Across Different Models.}
In addition to dataset-level statistics, we further compared the runtime efficiency of different topic modeling approaches on the \textbf{AI-DM} dataset. Figure \ref{fig:ai_dm_time} reports the average processing time (in minutes) required by various baseline methods and by \textbf{SciTopic}. Traditional probabilistic models such as \textbf{LDA} achieved the fastest runtime (3.75 minutes), but at the cost of weaker topic quality. Neural-based models like \textbf{ETM} and \textbf{NSTM} consumed substantially more time (49.99 and 154.44 minutes, respectively), while \textbf{ECRTM} incurred the heaviest computational burden (270.84 minutes). y comparison, \textbf{SciTopic} required 128.42 minutes. Although slower than lightweight baselines, its runtime remains moderate relative to other neural methods, and the performance improvements in coherence, diversity, and interpretability significantly outweigh the additional cost. This highlights the favorable balance between efficiency and quality, making \textbf{SciTopic} both scalable and practical for large-scale topic discovery.

\section{Conclusion}
\vspace{-1mm}
In this study, we propose an advanced topic modeling framework, \textbf{SciTopic}, which leverages LLMs to enhance the identification of topic structures in scientific texts. 
The core of this framework is refining document embeddings with entropy-based sampling techniques and the prompt-based triplet task, which refines the topic clustering process. 
Unlike traditional methods, our model does not rely on dimensionality reduction techniques, which results in better topic identification performance. 
Our experimental results indicate that \textbf{SciTopic} outperforms several baseline models, particularly TC and TD metrics, surpassing well-known models such as LDA, NMF, and BERTopic. Additionally, the incorporation of triplet tasks during the embedding refinement process offers deeper insights into topic relationships, while the class-based TF-IDF method further enriches topic representations. We validate the effectiveness of the framework using datasets from top conferences in artificial intelligence and data mining, demonstrating its superior performance in handling complex topic dynamics. Looking ahead, \textbf{SciTopic} holds significant potential for broader applications across various research domains, which provides a scalable tool for managing the surge in scientific publications. Future research may focus on integrating more diverse datasets and the enhancement of clustering interpretability, supporting more comprehensive trend analysis and more effective knowledge discovery.

\section*{Acknowledgment}

This work was supported by the National Natural Science Foundation of China (Grant Nos. 62406306 and 92470204) and the National Key Research and Development Program of China Grant (No. 2024YFF0729201).

\bibliography{custom}
\bibliographystyle{IEEEtran}


\end{document}